\documentclass{article}

\usepackage{microtype}
\usepackage{graphicx}
\usepackage{subfigure}
\usepackage{booktabs} %

\usepackage{amsmath,amsfonts,bm}

\def\eqref#1{equation~\ref{#1}}

\def\1{\bm{1}}

\DeclareMathAlphabet{\mathsfit}{\encodingdefault}{\sfdefault}{m}{sl}
\SetMathAlphabet{\mathsfit}{bold}{\encodingdefault}{\sfdefault}{bx}{n}

\usepackage{color}
\usepackage{hyperref}
\usepackage{url}
\usepackage{xspace}
\usepackage{adjustbox}
\usepackage{float}
\usepackage{tabularx}
\usepackage{multirow}
\usepackage{graphicx}
\usepackage{listings} %
\usepackage{newfloat}
\usepackage{inconsolata}
\usepackage[font=small,labelfont=bf]{caption}

\usepackage{sourcecodepro}

\DeclareFloatingEnvironment{listing}
\newcommand{\widelistingwidth}{0.85\textwidth}

\usepackage[accepted]{icml2019}

\icmltitlerunning{SmartChoices: Hybridizing Programming and Machine Learning \hfill \thepage}

\begin{document}

\twocolumn[
\icmltitle{SmartChoices: Hybridizing Programming and Machine Learning}

\icmlsetsymbol{equal}{*}

\begin{icmlauthorlist}
\icmlauthor{Victor Carbune}{goo}
\icmlauthor{Thierry Coppey}{goo}
\icmlauthor{Alexander Daryin}{goo}
\icmlauthor{Thomas Deselaers}{goo}
\icmlauthor{Nikhil Sarda}{goo}
\icmlauthor{Jay Yagnik}{goo}
\end{icmlauthorlist}

\icmlaffiliation{goo}{Google Research}

\icmlcorrespondingauthor{Victor Carbune}{vcarbune@google.com}

\icmlkeywords{Machine Learning, ICML}

\vskip 0.3in
]

\printAffiliationsAndNotice{}  %

\newcommand{\fix}[1]{\marginpar{\textcolor{blue}{FIX}}\textcolor{blue}{#1}}
\newcommand{\new}{\marginpar{NEW}}
\newcommand{\todo}[1]{\textcolor{red}{TODO: #1}}
\newcommand{\na}{--}
\newcommand{\widefigurewidth}{0.9\textwidth}

\definecolor{dkgreen}{rgb}{0,0.6,0}
\definecolor{gray}{rgb}{0.5,0.5,0.5}
\definecolor{mauve}{rgb}{0.58,0,0.82}

\newcommand{\SmartChoice}{SmartChoice\xspace}
\newcommand{\SmartChoices}{SmartChoices\xspace}

\lstset{frame=tb,
  language=Python,
  aboveskip=0mm, belowskip=0mm, lineskip=-1pt,
  emph={pvar,feedback,observe,predict,PVar,GOOD,BAD, Predict, Observe, Feedback},
  emphstyle=\textbf,
  showstringspaces=false,
  columns=fixed,
  basicstyle={\footnotesize\ttfamily},
  numbers=none,
  numberstyle=\tiny\color{gray},
  keywordstyle=\color{blue},
  commentstyle=\color{dkgreen},
  escapechar=|,
  stringstyle=\color{mauve},
  breaklines=true,
  breakatwhitespace=true,
  tabsize=3,
  showlines=true
}

\setlength{\abovecaptionskip}{0pt}
\setlength{\belowcaptionskip}{0pt}
\setlength{\topsep}{0pt}
\setlength{\parskip}{.5ex}
\renewcommand{\floatsep}{1ex}
\renewcommand{\textfloatsep}{1ex}
\renewcommand{\dblfloatsep}{1ex}
\renewcommand{\dbltextfloatsep}{1ex}

\begin{abstract}
We present \emph{SmartChoices}, an approach to making machine learning (ML) a first class citizen in programming languages which we see as one way to lower the entrance cost to applying ML to problems in new domains.
There is a growing divide in approaches to building systems: 
on the one hand, programming leverages human experts to define a system while on the other hand behavior is learned from data in machine learning. 
We propose to hybridize these two by providing a 3-call API which we expose through an object called \SmartChoice. 
We describe the \SmartChoices-interface, how it can be used in programming with minimal code changes, and demonstrate that it is an easy to use but still powerful tool by demonstrating improvements over not using ML at all on three algorithmic problems: binary search, QuickSort, and caches.
In these three examples, we replace the commonly used heuristics with an ML model entirely encapsulated within a \SmartChoice and thus requiring minimal code changes.
As opposed to previous work applying ML to algorithmic problems, our proposed approach does not require to drop existing implementations but 
seamlessly integrates into the standard software development workflow and gives full control to the software developer over how ML methods are applied. 
Our implementation relies on standard Reinforcement Learning (RL) methods. To learn faster, we use the heuristic function, which they are replacing, as an \emph{initial function}. We show how this initial function can be used to speed up and stabilize learning while providing a safety net that prevents performance to become substantially worse -- allowing for a safe deployment in critical applications in real life.
\end{abstract}

\section{Introduction}

Machine Learning (ML) has had many successes in the past decade in terms of techniques and systems as well as in the number of areas in which it is successfully applied. However, using ML has some cost that comes from the additional complexity added to software systems~\citep{sculley2014}. 
There is a fundamental impedance mismatch between the approaches to system building.
Software systems have evolved from the idea that experts have full control over the behavior of the system and specify the exact steps to be followed.
ML on the other hand has evolved from learning behavior by observing data. It allows for learning more complex but implicit programs leading to a loss of control for programmers since the behavior is now controlled by data.
We believe it is very difficult to move from one to another of these approaches, but that a hybrid between them needs to exist which
allows to leverage both the developer's domain-specific knowledge and the adaptability of ML systems.

We propose to hybridize ML with programming. We expose a new object called \SmartChoice exposing a 3-call API which is backed by ML-models and determines its value at runtime.
A developer will be able to use a \SmartChoice just like any other object, combine it with heuristics, domain specific knowledge, problem constraints, etc.\ in ways that are fully under the developer's control. 
This represents an \emph{inversion of control} compared to how ML systems are usually built.
\SmartChoices allow to integrate ML tightly into systems and algorithms whereas traditional ML systems are
built around the model.

Our approach combines methods from reinforcement learning (RL), online learning, with a novel API and aims to make using ML in software development easier by avoiding the overhead of going through the traditional steps of building an ML system:
(1) collecting and preparing training data,
(2) defining a training loss,
(3) training an initial model,
(4) tweaking and optimizing the model,
(5) integrating the model into their system, and
(6) continuously updating and improving the model to adjust for drift in the distribution of the data processed.

We show how these properties allow for applying ML in domains that have traditionally not been using it and that this is possible with minimal code changes.
We demonstrate that ML can help improve the performance of ``classical'' algorithms that typically rely on a heuristic.
The concrete implementation of \SmartChoices in this paper is based on standard deep RL. We emphasize that this is just one possible implementation.

In this paper we show \SmartChoices in the context of the Python programming language (PL) using concepts from object oriented PLs. The same ideas can be transferred directly to functional or imperative PLs, where a \SmartChoice could be modelled after a function or a variable.

We show how \SmartChoices can be used in three algorithmic problems -- binary search, QuickSort, and caches -- to improve performance by replacing the commonly used heuristic with an ML model with minimal code changes, leaving the structure of the original code (including potential domain-specific knowledge) untouched. We chose these problems as first applications for ease of reproducibility but believe that this demonstrates that our approach could benefit a wide range of applications, e.g. systems-applications, content recommendations, or modelling of user behavior. 

Further, we show how to use the heuristics that are replaced as ``\emph{initial functions}''  as means to guide the initial learning, help targeted exploration, and as a safety net to prevent very bad performance.

The main contributions of this paper are: 
(i) we propose a way to integrate ML methods directly into the software development workflow using a novel API;
(ii) we show how standard RL and online learning methods can be leveraged through our proposed API;
(iii) we demonstrate that this combination of ideas is simple to use yet powerful enough to improve performance of standard algorithms over not using ML at all.

\section{Software Development with \SmartChoices}
\label{sec:smartchoices}

A \SmartChoice has a simple API that allows the developer to provide enough information about its context, predict its value, and provide feedback about the quality of its predictions. \SmartChoices invert the control compared to common ML approaches that are model centric. Here, the developer has full control over how data and feedback are provided to the model, how inference is called, and how predictions are used.

To create a \SmartChoice, the developer chooses its output type (float, int, category, ...), shape, and range; defines which data the \SmartChoice is able to observe (type, shape, range); and optionally provides an initial function. In the following example we instantiate a scalar float \SmartChoice taking on values between $0$ and $1$, which can observe three scalar floats (each in the range between $0$ and $10$), and which uses a simple  initial function:
\begin{adjustbox}{width=\linewidth}
\begin{lstlisting}[frame=]
choice = SmartChoice(
  output_def=(float,shape=[1],range=[0,1]), 
  observation_defs={'low':(float,[1],[0,10]),
                    'high':(float,[1],[0,10]),
                    'target':(float,[1],[0,10])},
  initial_function=lambda observations:0.5)
\end{lstlisting}\vspace{-1ex}
\end{adjustbox}
The \SmartChoice can then be used. It determines its value when read using inference in the underlying ML model, e.g.\
\begin{lstlisting}[frame=, aboveskip=0.2ex, belowskip=0ex]
value = choice.Predict()
\end{lstlisting}\vspace{-1ex}
Specifically, developers should be able to use a \SmartChoice instead of a heuristic or an arbitrarily chosen constant. \SmartChoices can also take the form of a stochastic variable, shielding the developer from the underlying complexity of inference, sampling, and explore/exploit strategies.

The \SmartChoice determines its value on the basis of observations about the context that the developer passes in:
\begin{adjustbox}{width=\linewidth}
\begin{lstlisting}[frame=, aboveskip=0.2ex, belowskip=0ex]
choice.Observe('low', 0.12)
choice.Observe({'high':0.56,'target':0.43})
\end{lstlisting}\vspace{-1ex}
\end{adjustbox}
A developer might provide additional side-information into the \SmartChoice that an engineered heuristic would not be using but which a powerful model is able to use in order to improve performance.

The developer provides feedback about the quality of previous predictions once it becomes available:
\begin{lstlisting}[frame=, aboveskip=0.2ex, belowskip=0ex]
choice.Feedback(reward=10)
\end{lstlisting}\vspace{-1ex}

In this example we provide numerical feedback. Following common RL practice a \SmartChoice aims to maximize the sum of reward values received over time (possibly discounted). In other setups, we might become aware of the correct value in hindsight and provide the ``ground truth'' answer as feedback, turning the learning task into a supervised learning problem. Some problems might have multiple metrics to optimize for (run time, memory, network bandwidth) and the developer might want to give feedback for each dimension. 

This API allows for integrating \SmartChoices easily and transparently into existing applications with little overhead. See listing~\ref{lst:bsearch_sc} for how to use the \SmartChoice created above in binary search. In addition to the API calls described above, model hyperparameters can be specified through additional configuration, which can be tuned independently. The definition of the \SmartChoice only determines its interface (i.e.\ the types and shapes of inputs and outputs).

\section{Initial Functions in SmartChoices}
\label{sec:initial}

We allow 
for the developer to pass an initial function to the \SmartChoice. We anticipate that in many cases the initial function will be the heuristic that the \SmartChoice is replacing. Ideally it is a reasonable guess at what values would be good for the \SmartChoice to return. %
The \SmartChoice will use this initial function to avoid bad performance in the initial predictions, 
and observe the behavior of the initial function to guide its own learning process, similar to imitation learning~\citep{imitationlearningsurvey}. 
The existence of the initial function should strictly improve the performance of a \SmartChoice. In the worst case, the \SmartChoice could choose to ignore it completely, but ideally it will allow the \SmartChoice to explore solutions which are not easily reachable from a random starting point.
Further, the initial function plays the role of a heuristic policy which explores the state and action space generating initial trajectories which are then used for learning. 
Even though such exploration is biased, off-policy RL can train on this data.
In contrast to imitation learning where an agent tries to become as good as the expert, we explicitly aim to outperform the initial function as quickly as possible, similar to \cite{Schmitt2018KickstartingDR}.

For a \SmartChoice to make use of the initial heuristic, and to balance between learning a good policy and the safety of the initial function, it relies on a \emph{policy selection strategy}.  This strategy switches between
exploiting the learned policy, exploring alternative values, and using the initial function.  It can be applied at the action or episode level depending on the requirements.
Finally, the initial function provides a safety net: in case the learned policy starts to misbehave, the \SmartChoice can always fallback to the initial function with little cost.

\section{SmartChoices in Algorithms}
\label{sec:applications}
In this section, we describe how \SmartChoices can be used in three different algorithmic problems and how a developer can leverage the power of machine learning easily with just a few lines of code. We show experimentally how using \SmartChoices helps improving the algorithm performance. 
The interface described above naturally translates into an RL setting: 
the inputs to \lstinline{Observe} calls are combined into the state, 
the output of the \lstinline{Predict} call is the action, 
and \lstinline{Feedback} is the reward. 

To evaluate the impact of \SmartChoices we measure \textbf{cumulative regret} over training episodes.
Regret measures how much worse (or better when it is negative) a method performs compared to another method. Cumulative regret captures whether a method is better than another method over all previous decisions.
For practical use cases we are interested in two properties: 
(1) Regret should never be very high to guarantee acceptable performance of the \SmartChoice under all circumstances.
(2) Cumulative regret should become permanently negative as early as possible. This corresponds to the desire to have better performance than the baseline model as soon as possible.

Unlike the usual setting which distinguishes a training and evaluation mode, we perform evaluation from the point of view of the developer without this distinction. The developer just plugs in the \SmartChoice and starts running the program as usual. Due to the online learning setup in which \SmartChoices are operating, overfitting does not pose a concern~\citep{Dekel2005DataDrivenOT}.
The (cumulative) regret numbers thus do contain potential performance regressions due to exploration noise. This effect could be mitigated by performing only a fraction of the runs with exploration.

In our experiments we do not account for the computational costs of inference in the model. 
The goal of our study is to demonstrate that the proposed approach is generally feasible and that with minimal code changes ML can be used in programming. 
While for algorithms, like those we are experimenting with here, the actual run time does matter we believe that advances in specialized hardware will enable running machine learning models at insignificant cost~\citep{Kraska2018TheCF}.
Further, even if such cost seem high, we see \SmartChoices applicable to a wide variety of problems: e.g. relying on expensive approximation heuristics or working with inherently slow hardware, such as filesystems where the inference time is less relevant.
And lastly, our approach is applicable to a wide variety of problems ranging from systems problems, over user modelling, to content recommendation where the computational overhead for ML is not as problematic.

\begin{figure}[tb]
\begin{center}
\includegraphics[width=.9\linewidth,trim={0.3cm 16.9cm 11.6cm 0},clip]{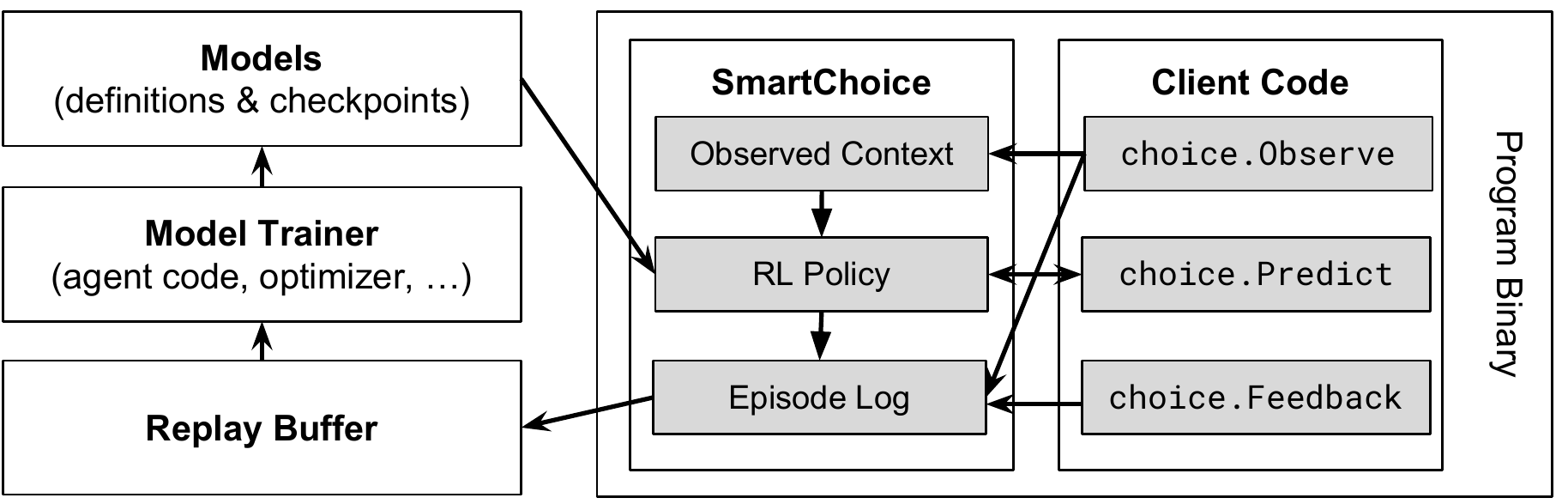}
\vspace{-3ex}
\caption{An overview of the architecture for our experiments how client code communicates with a \SmartChoice and how the model for the \SmartChoice is trained and updated.}
\label{fig:architectures}
\end{center}
\end{figure}

Our implementation  currently is a small library exposing the \SmartChoice interface to client applications~(fig.~\ref{fig:architectures}). 
A \SmartChoice assembles observations, actions, and feedback into episode logs that are passed to a replay buffer. The models are trained asynchronously. When a new checkpoint becomes available the \SmartChoice loads it for use in consecutive steps.

\subsection{Experiment Setup}
\label{sec:models}
To enable \SmartChoices we leverage recent progress in RL for modelling and training. It allows to apply \SmartChoices to the most general use cases. While we are only looking at RL methods here, \SmartChoices could be used with other learning methods such as multi-armed bandits or supervised learning.
We are building our models on DDQN \citep{hasselt:aaai2016-ddqn} for categorical outputs and on  TD3~\citep{fujimoto:icml2018-td3} for continuous outputs. 
DDQN is a de facto standard in RL since its success in AlphaGo~\citep{alphago}. TD3 is a recent modification to DDPG~\citep{lilicrap:ddpg2015} using a second critic network to avoid overestimating the expected reward.
We summarize the hyperparameters used in our experiments in (table~\ref{tab:settings}).

\begin{table}[tb]
  \centering
  \newcommand{\yes}{yes}
  \newcommand{\no}{no}
 \caption{Parameters for the different experiments described below
 (FC=fully connected layer, LR=learning rate). See \citep{deep_rl_that_matters} for details on these parameters.}
 \label{tab:settings}
  \begin{adjustbox}{width=.48\textwidth}
  \begin{tabular}{|l|c|c|c|c|}
    \hline
                   & Binary search & QuickSort & Caches (discrete) & Caches (continuous) \\\hline
    Learning algorithm      &  TD3 & DDQN & DDQN & TD3 \\\hline
    Actor network           & $\text{FC}_{16} \to \tanh$ & --  & -- & $\text{FC}_{10} \rightarrow \tanh$ \\\hline
    Critic/value network          & $\text{FC}_{16}$ & $(\text{FC}_{16}, \text{ReLU})^2 \to \text{FC}$ & $(\text{FC}_{10}, \text{ReLU})^2 \to \text{FC}$ & $\text{FC}_{10}$ \\\hline
    Key embedding size      & -- & -- & \multicolumn{2}{|c|}{8}\\\hline
    Discount                & $0.8$, $0$ & $0$ & \multicolumn{2}{|c|}{$0.8$} \\\hline
    LR actor                & $10^{-3}$ & -- & -- & $10^{-4}$ \\\hline
    Initial function decay  & \yes & \multicolumn{3}{|c|}{\no} \\\hline
    Batch size              & \multicolumn{2}{|c|}{256} & \multicolumn{2}{|c|}{1024} \\\hline
    Action noise $\sigma$   & $0.03$ & -- & -- & $0.01$ \\\hline
    Target noise $\sigma$   & $0.2$  & -- & -- & $0.01$ \\\hline
    Temperature             & -- & \multicolumn{2}{|c|}{$0.1$} & -- \\\hline
    Update ratio ($\tau$)   & 0.05 & \multicolumn{3}{|c|}{$0.001$}\\\hline
    \multicolumn{5}{|l|}{Common: Optimizer: Adam; 
                                            LR critic: $10^{-4}$;
                                            Replay buffer: Uniform, FIFO, size 20000;
                                            Update period: 1.
                                            }\\\hline
  \end{tabular}
  \end{adjustbox}
\end{table}

While these hyperparameters are now new parameters that the developer can tweak, we hypothesize that on the one hand tuning hyperparameters is often simpler than manually defining new problem-specific heuristics, and on the other hand that improvements on automatic model tuning from the general machine learning community will be easily applicable here too.

Our policy selection strategy starts by only evaluating the initial function and then gradually starts to increase the use of the learned policy. It keeps track of the received rewards of these policies adjusts the use of the learned policy depending on its performance.
We show the usage rate of the initial function when we use it (fig.~\ref{fig:bsearch_results}, bottom) demonstrating the effectiveness of this strategy.

\subsection{Binary Search}
\label{sec:bsearch}

\begin{listing*}[tb]
\caption{Standard binary search (left) and a simple way to use a \SmartChoice in binary search (right).\label{lst:bsearch_sc}}
\vspace{1ex}
\centering
\begin{adjustbox}{width=\widelistingwidth}
\begin{minipage}{.5\textwidth}
\begin{lstlisting}[numbers=left]
def bsearch(x, a, l=0, r=len(a)-1):
  if l > r: return None


  q = 0.5
  m = int(q*l + (1-q)*r)
  if a[m] == x:
    return m

  if a[m] < x:
    return bsearch(x, a, m+1, r)
  return bsearch(x, a, l, m-1)
\end{lstlisting}
\end{minipage}
~
\begin{minipage}{.5\textwidth}
\begin{lstlisting}[numbers=right]
def bsearch(x, a, l=0, r=len(a)-1):
  if l > r: return None
  choice.Observe({'target':x,
    'low':a[l], 'high':a[r]})
  q = choice.Predict()
  m = int(q*l + (1-q)*r)
  if a[m] == x:
    return m
  choice.Feedback(-1)
  if a[m] < x:
    return bsearch(x, a, m+1, r)
  return bsearch(x, a, l, m-1)
\end{lstlisting}
\end{minipage}
\end{adjustbox}
\end{listing*}

Binary search \citep{binarysearch} is a standard algorithm for finding the location $l_x$ of a target value $x$ in a sorted array $A = \{a_0, a_1, \dots, a_{N-1} \}$ of size $N$.
Binary search has a worst case runtime complexity of $\lceil\log_2(N)\rceil$ steps when no further knowledge about the distribution of data is available.
Prior knowledge of the data distribution can help reduce the average runtime: e.g. in case of an
uniform distribution, the location of $x$ can be approximated using linear interpolation $l_x \approx (N - 1) {(x - a_0)}/{(a_{N-1} - a_0)}$. 
We show how \SmartChoices can be used to speed up binary search by learning to estimate the position $l_x$ for a more general case.

The \emph{simplest way} of using a \SmartChoice is to directly estimate the location $l_x$ and incentivize the search to do so in as few steps as possible by penalizing each step by the same negative reward (listing~\ref{lst:bsearch_sc}). At each step, the \SmartChoice observes the values $a_L$, $a_R$ at both ends of the search interval and the target $x$. The \SmartChoice output $q$ is used as the relative position of the next read index $m$, such that $m = q L + (1 - q) R$.

In order to give a \emph{stronger learning signal} to the model, the developer can incorporate problem-specific knowledge into the reward function or into how the \SmartChoice is used. One way to \emph{shape the reward} is to account for problem reduction.
For binary search, reducing 
the size of the remaining search space will speed up the search proportionally and should be rewarded accordingly. By replacing the step-counting reward in listing~\ref{lst:bsearch_sc} (line 9) with the search range reduction $(R_t-L_t)/(R_{t+1}-L_{t+1})$, we directly reward reducing the size of the search space.
By shaping the reward like this, we are able to attribute the feedback signal to the current prediction and to reduce the problem from RL to contextual bandit (which we implement by using a discount factor of $0$).

Alternatively we can \emph{change the way the prediction is used} to cast the problem in a way that the \SmartChoice learns faster and is unable to predict very bad values. For many algorithms (including binary search) it is possible to predict a combination of (or choice among) several existing heuristics rather than predicting the value directly.
We use two heuristics:
(a) vanilla binary search which splits the search range $\{a_{L}, \ldots, a_{R}\}$ into two equally large parts using the split location $l^{v}=(L+R)/2$, and 
(b) interpolation search which interpolates the split location as $l^{i} = {((a_R - v) L + (v - a_L) R)}/{(a_R - a_L)}$.
We then use the value $q$ of the \SmartChoice to mix between these heuristics to get the predicted split position $l^q = q l^{v} + (1-q) l^{i}$.
Since in practice both of these heuristics work well on many distributions, any point in between will also work well.
This reduces the risk for the \SmartChoice to pick a value that is really bad which in turn helps learning. 
A disadvantage is that it is impossible to find the optimal strategy if its values lie outside of the interval between $l^v$ and $l^i$.

To \emph{evaluate} our approaches we use a test environment where in each episode, we search a random element in a sorted array of $5000$ elements taken from a randomly chosen distribution (uniform, triangular, normal, pareto, power, gamma and chisquare), with values in $[-10^4,10^4]$.

\begin{figure}[tb]
\begin{center}
\includegraphics[width=.95\linewidth]{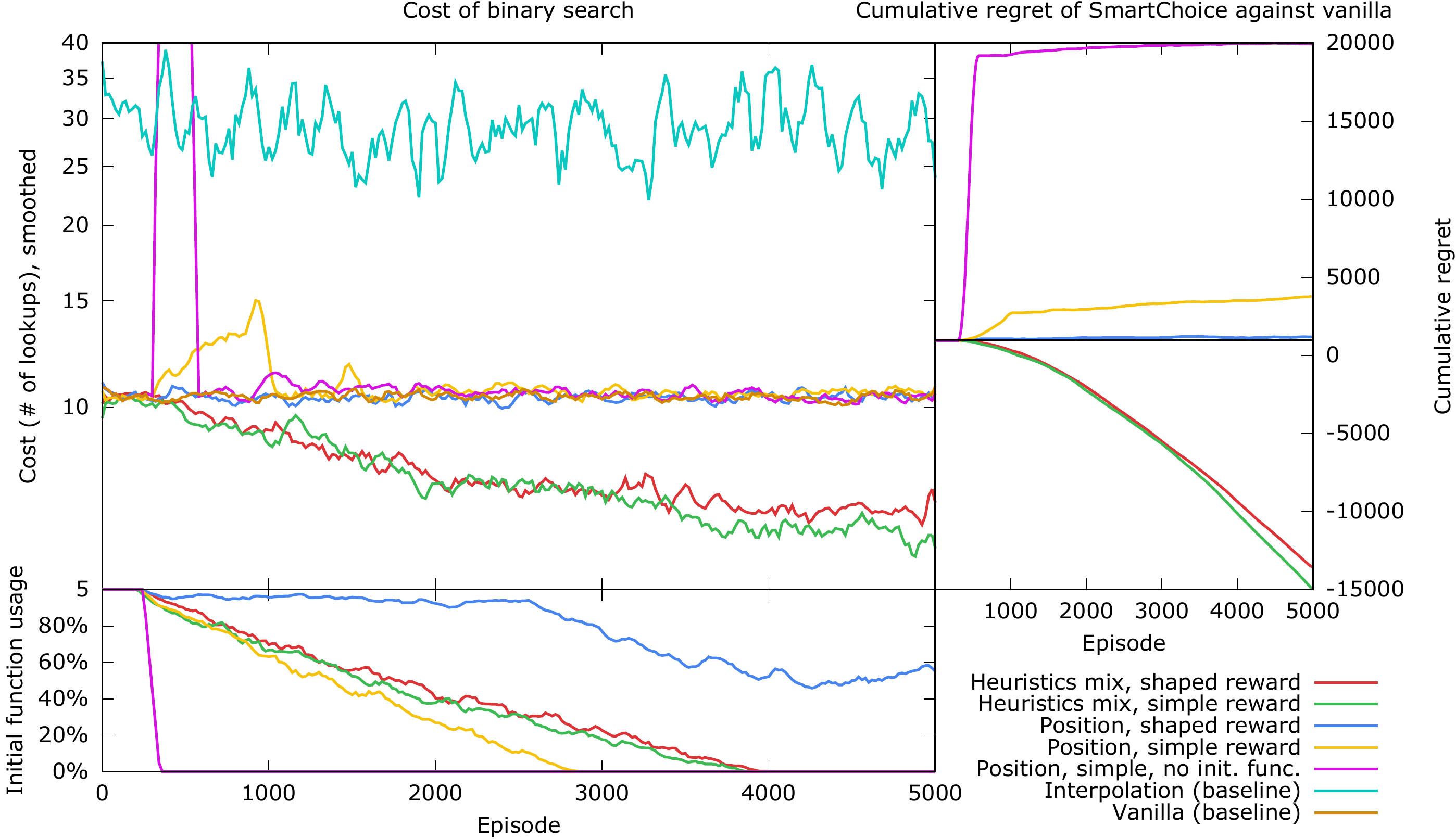}
\end{center}
\vspace{-3ex}
\caption{The cost of different variants of binary search (top left), cumulative regret compared to vanilla binary search (right), and initial function usage (bottom).}
\label{fig:bsearch_results}
\end{figure}

Figure~\ref{fig:bsearch_results} shows the results for the different variants of binary search using a \SmartChoice and compares them to the vanilla binary search baseline. The results show that the simplest case (pink line) where we directly predict the relative position with the simple reward and without using an initial function performs poorly initially but then becomes nearly as good as the baseline (cumulative regret becomes nearly constant after an initial bad period). 
The next case (yellow line) has an identical setup but we are using the initial function and we see that the initial regret is substantially smaller. 
By using the shaped reward (blue line), the \SmartChoice is able to learn the behavior of the baseline quickly.
Both approaches that are mixing the heuristics (green and red lines) significantly outperform the baselines.

\subsection{QuickSort}
\label{sec:qsort}
QuickSort \citep{QuickSort} sorts an array in-place by partitioning it into two sets (smaller/larger than the pivot) recursively until the array is fully sorted. QuickSort is one of the most commonly used sorting algorithms where many heuristics have been proposed to choose the pivot element.
While the average time complexity of QuickSort is $\theta(N \log(N))$, a worst case time complexity of $O(N^2)$ can happen when the pivot elements are badly chosen. The optimal choice for a pivot is the median of the range, which splits it into two parts of equal size.

\begin{listing*}[tb]
\caption{A QuickSort implementation that uses a \SmartChoice to choose the number of samples to compute the next pivot. As feedback, we use the cost of the step compared to the optimal partitioning.}
\label{lst:qsort_pvar}
\vspace{1ex}
\centering
\begin{adjustbox}{width=\widelistingwidth}
\begin{minipage}{.45\textwidth}
\begin{lstlisting}[numbers=left]
def qsort(a, l=0, r=len(a)):
  if r <= l+1: 
    return
  m = pivot(a, l, r)
  qsort(a, l, m-1)
  qsort(a, m+1, r)

def delta_cost(c_pivot, n, a, b):
  # See eq.|\textcolor{dkgreen}{~\ref{eq:delta_cost}}|
\end{lstlisting}
\end{minipage}
\begin{minipage}{.55\textwidth}
\begin{lstlisting}[numbers=right]
def pivot(a, l, r):
  choice.Observe({'left':l, 'right':r})
  q = min(1+2*choice.Predict(), r-l)
  v = median(sample(a[l:r], q))
  m = partition(a, l, r, v)
  c = cost_of_median_and_partition()
  d = delta_cost(c, r-l, m-l, r-m)
  choice.Feedback(1/d)
  return m
\end{lstlisting}
\end{minipage}
\end{adjustbox}
\end{listing*}

To improve QuickSort using a \SmartChoice we aim at tuning the pivot selection heuristic. 
To allow for sorting arbitrary types, we use the \SmartChoice to determine the number of random samples to pick from the array to sort, and use their median as the partitioning pivot (listing~\ref{lst:qsort_pvar}).
As \emph{feedback signal} for a recursion step, we estimate the impact of the pivot selection on the computational cost $\Delta c$.
\begin{equation}
\label{eq:delta_cost}
\Delta c = \frac{c_{\text{piv}} + \Delta c_{\text{rec}}}{c_{\text{expected}}}
              = \frac{c_\text{piv} + (a \log a + b \log b - 2 \frac{n}{2} \log\frac{n}{2})}{n \log n},
\end{equation}
where $n$ is the size of the array, $a$ and $b$ are the sizes of the partitions with $n=a+b$ and $c_{\text{piv}} = c_\text{median} + c_\text{partition}$ is the cost to compute the median of the samples and to partition the array. $\Delta c_{\text{rec}}$ takes into account how close the current partition is to the ideal case (median). The cost is a weighted sum of number of reads, writes, and comparisons. Similar to the shaped reward in binary search, this reward allows us to reduce the RL problem to a contextual bandit problem and we use a discount of $0$. 

\begin{figure}[tb]
\begin{center}
\includegraphics[width=.95\linewidth]{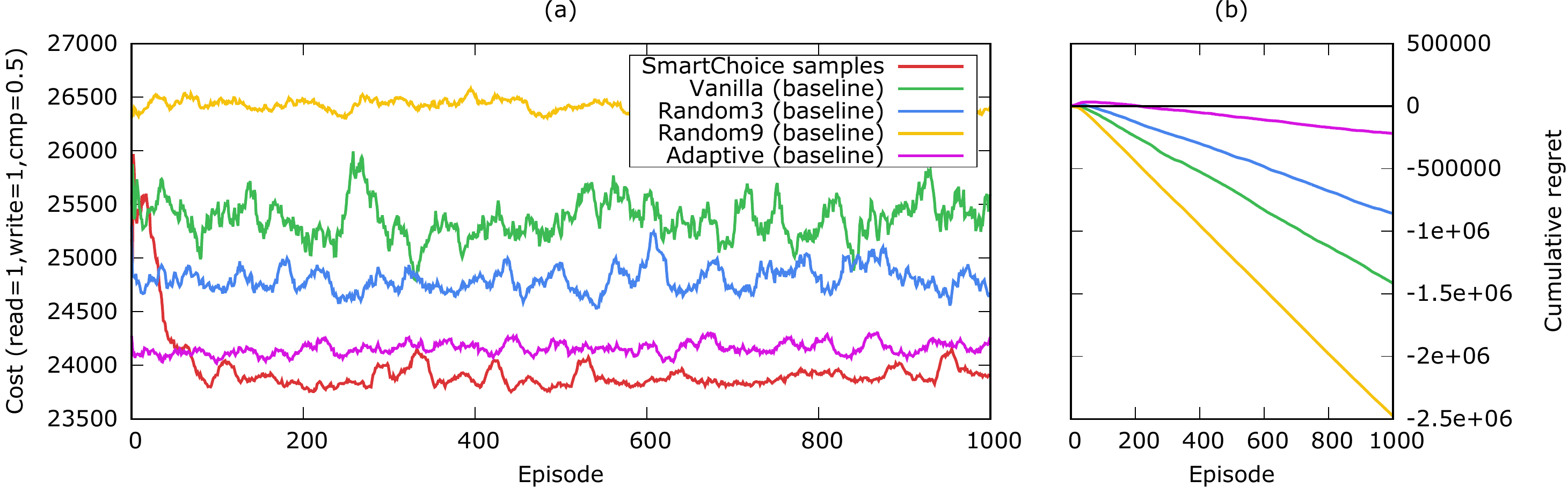}
\end{center}
\vspace{-3ex}
\caption{Results from using a \SmartChoice for selecting the number of pivots in QuickSort. (a) shows the overall cost for the different baseline methods and for the variant with a \SmartChoice over training episodes. (b) shows the cumulative regret of the \SmartChoice method compared to each of the baselines over training episodes.}
\label{fig:qsort_results}
\begin{center}
\includegraphics[trim={2cm 13.1cm 2cm 2.1cm},clip,width=.95\linewidth]{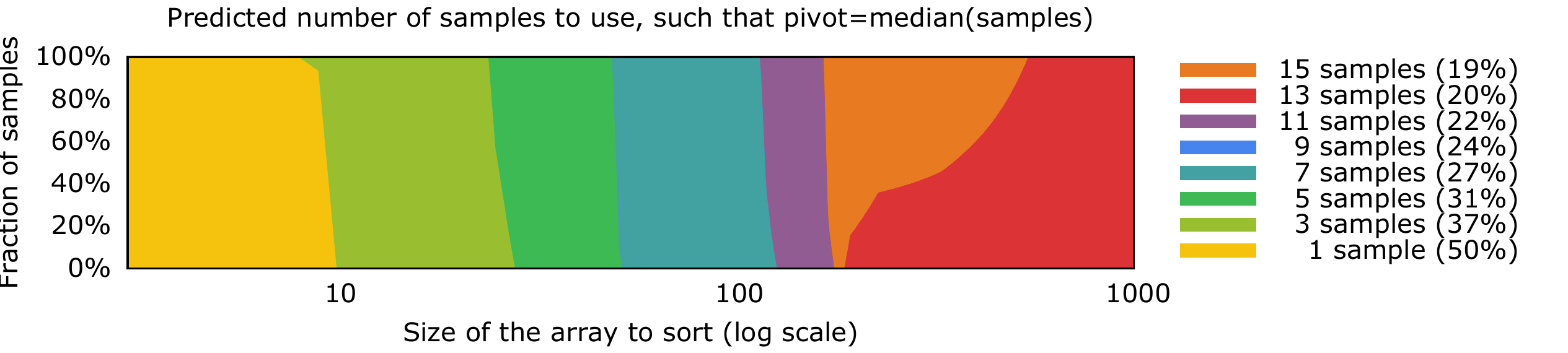}
\end{center}
\vspace{-3ex}
\caption{Number of pivots chosen by the \SmartChoice in QuickSort after 5000 episodes. The expected approximation error of the median is given in the legend, next to the number of samples.}
\label{fig:qsort_output}
\end{figure}

For \emph{evaluation} we are using a test environment where we sort randomly shuffled arrays.
Results of the experiments are presented in fig.~\ref{fig:qsort_results} and show that the learned method outperforms all baseline heuristics within less than 100 episodes. `Vanilla' corresponds to a standard QuickSort implementation that picks one pivot at random in each step. `Random3' and `Random9' sample 3 and 9 random elements respectively and use the median of these as pivots. `Adaptive' uses the median of $\max(1, \lfloor \log_2(n) - 1 \rfloor)$ randomly sampled elements as pivot  when partitioning a range of size $n$. It uses more samples at for larger arrays, leading to a better approximation of the median, and thus to faster problem size reduction.

Fig.~\ref{fig:qsort_output} shows that the \emph{\SmartChoice learns a non-trivial policy}. 
The \SmartChoice learns to select more samples at larger array sizes which is similar to the behavior that we hand-coded in the adaptive baseline but in this case no manual heuristic engineering was necessary and a better policy was learned. Also, note that a \SmartChoice-based method is able to adapt to changing environments which is not the case for engineered heuristics.
One surprising result is that the \SmartChoice prefers 13 over 15 samples at large array sizes. We hypothesize this happens because relatively few examples of large arrays are seen during training (one per episode, while arrays of smaller sizes are seen multiple times per episode).

\subsection{Caches}
\label{sec:caches}
Caches are a commonly used component to speed up computing systems.
They use a \emph{cache replacement policy} (CRP) to determine which element to evict when the cache is full and a new element needs to be stored. 
Probably the most popular CRP is the \emph{least recently used} (LRU) heuristic which evicts the element with the oldest access timestamp. A number of approaches have been proposed to improve cache performance using machine learning (see sec.~\ref{sec:relwork}).
We propose two different approaches how \SmartChoices can be used in a CRP to improve cache performance.

\begin{listing*}[tb]
\caption{Cache replacement policy directly predicting eviction decisions (\emph{Discrete}).}
\label{lst:discrete_cache}
\vspace{1ex}
\centering
\begin{adjustbox}{width=\widelistingwidth}
\begin{minipage}{.5\textwidth}
\begin{lstlisting}[numbers=left]
keys = ...  # keys now in cache.

# Returns evicted key or None.
def miss(key):
  choice.Feedback(-1) # Miss penalty.
  choice.Observe('access', key)
  choice.Observe('memory', keys)
  return evict(choice.Predict())
\end{lstlisting}
\end{minipage}
\begin{minipage}{.5\textwidth}
\begin{lstlisting}[numbers=right]
def evict(i):
  if i >= len(keys): return None
  choice.Feedback(-1) # Evict penalty.
  choice.Observe('evict', keys[i])
  return keys[i]
def hit(key):
  choice.Feedback(1) # Hit reward.
  choice.Observe('access', key)
\end{lstlisting}
\end{minipage}
\end{adjustbox}
\caption{Cache replacement policy using a priority queue (\emph{Continuous}).}
\label{lst:continuous_cache}
\vspace{1ex}
\centering
\begin{adjustbox}{width=\widelistingwidth}
\begin{minipage}{.5\textwidth}
\begin{lstlisting}[numbers=left]
q = min_priority_queue(capacity)
def priority(key):
  choice.Observe(...)
  score = choice.Predict()
  score *= capacity * scale
  return time() + score
\end{lstlisting}
\end{minipage}
\begin{minipage}{.5\textwidth}
\begin{lstlisting}[numbers=right]
def hit(key):
  choice.Feedback(1) # Hit reward.
  q.update(key, priority(key))
def miss(key):
  choice.Feedback(-1) # Miss penalty.
  return q.push(key, priority(key))
\end{lstlisting}
\end{minipage}
\end{adjustbox}
\end{listing*}

\textbf{Discrete} (listing~\ref{lst:discrete_cache}): A \SmartChoice directly predicts which element to evict
or chooses not to evict at all (by predicting an invalid index).
That is, the \SmartChoice learns to become a CRP itself. While this is the simplest way to use a \SmartChoice, it 
makes it more difficult to learn a CRP better than LRU (in fact, even learning to be on par with LRU
is non-trivial in this setting).

\textbf{Continuous} (listing~\ref{lst:continuous_cache}): A \SmartChoice is used to enhance LRU by predicting an offset to the last access timestamp. Here, the \SmartChoice learns which items to keep in the cache longer and which items to evict sooner.
In this case it becomes trivial to be as good as LRU by predicting a zero offset. The \SmartChoice value in $(-1, 1)$ is scaled to get a reasonable value range for the offsets. It is also possible to choose not to store the element by predicting a sufficiently negative score.

In both approaches the feedback given to the \SmartChoice is whether an item was found in the cache ($+1$) or not ($-1$). In the discrete approach we also give a reward of $-1$ if the eviction actually takes place.

In our implementation the observations are the history of accesses, memory contents, and evicted elements. The \SmartChoice can observe
(1) keys as a categorical input or (2) features of the keys.

\begin{figure}[tb]
\begin{center}
\includegraphics[width=.8\linewidth,trim={0.9cm 17cm 15.5cm 1.1cm},clip]{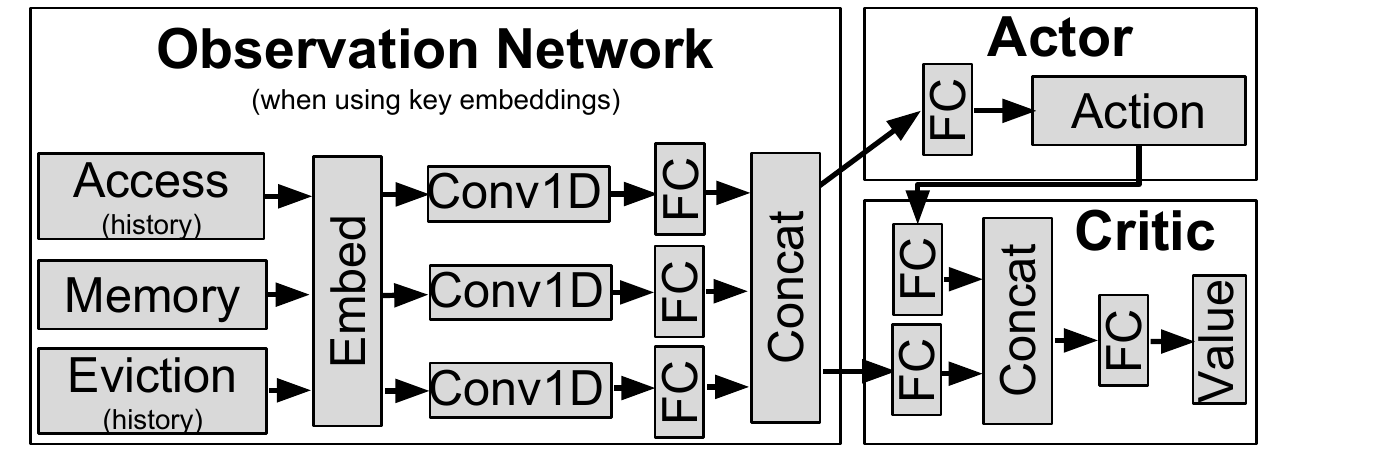}
\end{center}
\vspace{-3ex}
\caption{The architecture of the neural networks for TD3 with key embedding network.}
\label{fig:architectures-2}
\end{figure}

Observing \emph{keys as categorical input} allows to avoid feature engineering and enables directly
learning the properties of particular keys (e.g.\ which keys are accessed the most) but makes it difficult to deal with rare and unseen keys.
To handle keys as input we train an embedding layer shared between the actor and critic networks (fig.~\ref{fig:architectures-2}).

As \emph{features of the keys} we observe historical frequencies computed over a window of fixed size.
This approach requires more effort from the developer to implement such features, but pays off with better
performance and the fact that the model does not rely on particular key values.

\begin{figure}[tb]
\begin{center}
\includegraphics[width=.95\linewidth]{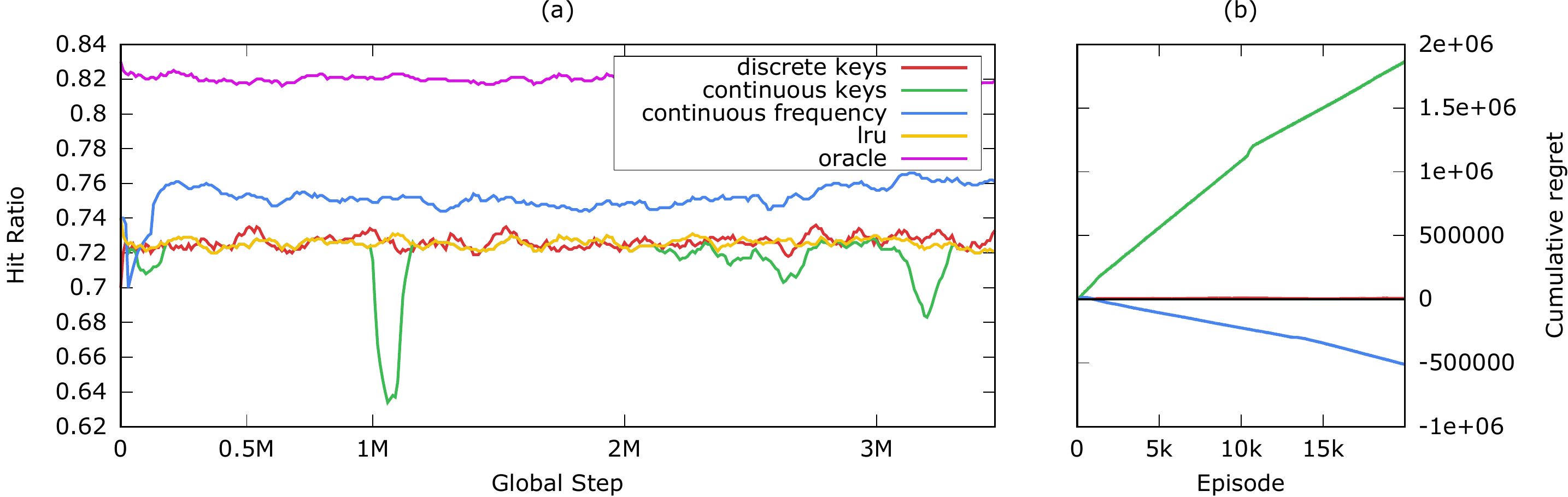}
\includegraphics[width=.95\linewidth]{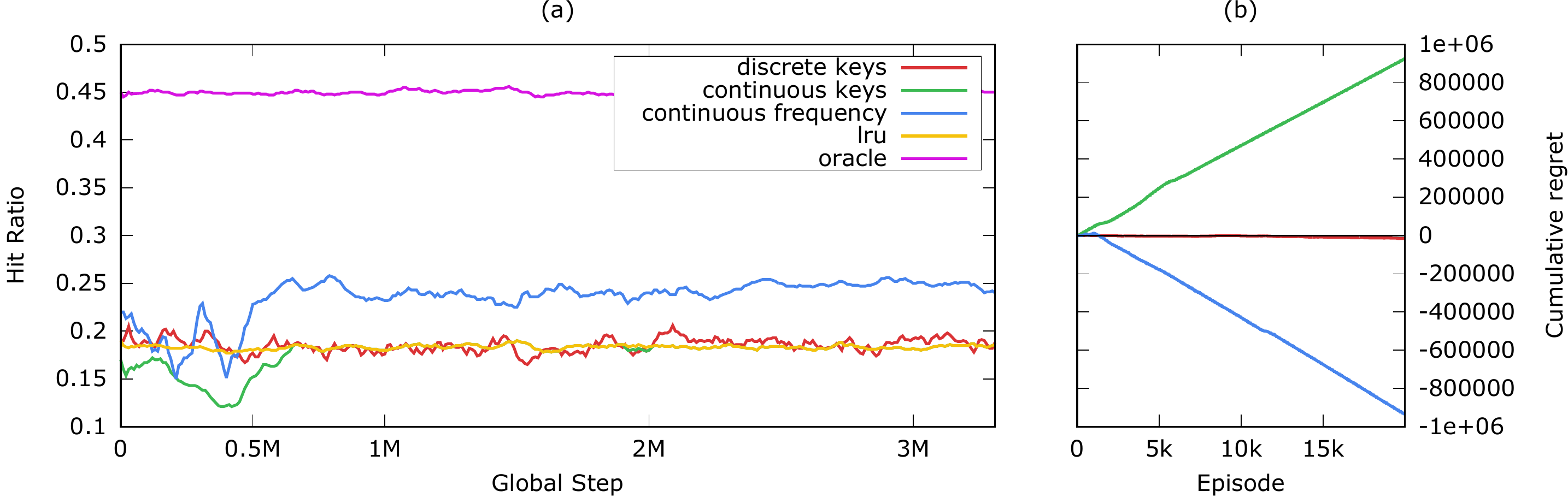}
\end{center}
\vspace{-3ex}
\caption{Cache performance for power law access patterns. Top: $\alpha=0.1$, bottom: $\alpha=0.5$. (a) Hit Ratio (w/o exploration) and (b) Cumulative Regret (with exploration)}
\label{fig:CacheHitRatio}
\end{figure}

We experiment with three combinations of these options: (1) discrete caches observing keys, (2) continuous caches observing keys, (3) continuous caches observing frequencies. For \emph{evaluation} we use a cache with size $10$ and integer keys from $1$ to $100$. 
We use two synthetic access patterns of length $1000$, sampled i.i.d.\ from a power law distribution with  $\alpha=0.1$ and $\alpha=0.5$. Fig.~\ref{fig:CacheHitRatio} shows results for the three variants of predicted caches, a standard LRU cache, and an oracle cache to give a theoretical, non-achievable, upper bound on the performance.

We look at the hit ratio without exploration to understand the potential performance of the model once learning has converged. However, cumulative regret is still reported under exploration noise.

Both implementations that work directly on key embeddings learn to behave similar to the LRU baseline without exploration (comparable hit ratio). However, the continuous variant pays a higher penalty for exploration (higher cumulative regret).
Note that this means that the continuous variant learned to predict constant offsets (which is trivial), however the discrete implementation actually learned to become an LRU CRP which is non-trivial. 
The continuous implementation with frequencies quickly outperforms the LRU baseline, making the cost/benefit worthwhile long-term (negative cumulative regret after a few hundred episodes).

\subsection{Reproducibility: Goals and Metrics}
\label{sec:reproducibility}

Nonetheless,  Similar to many works that build on RL technology, we are faced with the reproducibility issues described by \cite{deep_rl_that_matters}. Among multiple runs of any experiment, only some runs exhibit the desired behavior, which we report.  In the ``failing'' runs, we observe baseline performance because the initial function acts as a safety net. Thus, our experiments show that we can outperform the baseline heuristics without a high risk to fail badly. 
The design construct specific to \SmartChoices and what distinguishes it from standard Reinforcement Learning is that it is applied in software control where often developers are able to provide safe initial functions or write the algorithm in a way that limits the cost of a poorly performing policy.
While we do not claim to have the solution to address reproducibility, the use of the initial function can mitigate it and any solution to better reproducibility and higher stability  developed by the community will be applicable in our approach as well.

In table~\ref{tab:reproducibility}, we provide details on the reproducibility and performance of our experiments over 100 identical experiments for each of the problems described earlier. The table shows the cumulative regret and the break even point for our experiments for various quantiles and as the mean. Cumulative regret indicates how much worse our method is than not using ML at all -- if it's negative it means that it's better than not using it. The break even point is the number of episodes after which cumulative regret becomes negative and never positive anymore. In some experiments the break even point is not reached. We report the percentage of runs in which it was reached in the `mean' column. 

\begin{table*}[tb]
    \centering
    \small
    \caption{Reproducibility data for our experiments: We report cumulative regret for different quantiles of experiments at different training episodes as well as the average over all episodes. We also report the respective break even point as a number of episodes, which is the number of training episodes at which cumulative regret becomes negative and never positive anymore. For the break even point we report the percentage of runs in which the break even point was reached in the column ``mean''.}
        \label{tab:reproducibility}
    \begin{adjustbox}{width=\textwidth}
    
    \begin{tabular}{|l|l||r|r|r|r|r|r|r|r|r|r|}
    \hline
        Problem & Percentile            & 1      & 5     &    10 & 25    & 50    & 75    & 90    & 95   &    99  & mean \\ \hline
  \multirow{3}{*}{Binary Search (N=120)}              & Cum.\ Regret @5K episodes   & -2.71  & -2.66 & -2.62 & -2.45 & -2.03 & -1.01 & 0.44  & 0.70 &  0.78 &  -1.59\\ 
   &    Cum.\ Regret @50K episodes  & -3.99  & -3.83 & -3.76 & -3.64 & -3.34 & -2.85 & 3.80  & 3.86 &  3.92 & -2.20 \\ 
                & Break-even  (episodes)   & 127    & 201   & 271   & 417   & 758   & 2403  & $\infty$ & $\infty$ & $\infty$ & 85\% \\
    \hline
    \multirow{3}{*}{QuickSort (N=115)}            & Cum.\ Regret @1K episodes  & -1273 & -1248 & -1214 & -1146 & -1029 & -916  & -409  & 372 & 425 & -913\\
       & Cum.\ Regret @10K episodes & -1356 & -1306 & -1267 & -1219 & -1146 & -1034 & -945  & -285  & 393 & -1064\\ 
                & Break-even  (episodes)  & 0    & 0    & 0    & 37   & 93   & 141  & 307  & 7370 & $\infty$ & 94\% \\
    \hline
   \multirow{2}{*}{Cache (N=100)}        & Cum.\ Regret @20K episodes      & -8.25  & -5.88 & -3.49 & -0.00 & 0.00  & 0.02  & 0.34  & 0.84  & 2.17 & -0.52\\ 
                & Break even  (episodes)   & 32    & 157   & 472   & $\infty$   & $\infty$  & $\infty$  & $\infty$ & $\infty$ & $\infty$ & 26\% \\
    \hline
    \end{tabular}
  \end{adjustbox}
  
\end{table*}

We want to highlight that, while the experiments for some problems are more reproducible than others, our approach does not perform substantially worse than the initial function provided by the developer, e.g. cumulative regret for none of the problems grows very large, indicating that performance remains acceptable. This is very visible for the cache experiments: While only for 26\% of the runs the break even point was reached, meaning that the cache performs strictly better than before, it only performs worse than before in 14\% of the runs. For 60\% of the runs, the use of ML does neither help nor hurt compared to using the LRU heuristic.

\section{Related work}
\label{sec:relwork}

The most relevant work to our proposed interface is \citep{Chang2016ACA} where a programming interface is proposed for joint prediction and a method that allows for unifying the implementation for training and inference. 
Similarly, Probabilistic programming \citep{gordon2014probabilistic} introduces interfaces which simplify the developer complexity when working with statistical models and conditioning variable values on run-time observations. Our proposed interfaces are at a higher level in that the user does not need to know about the inner workings of the underlying models. In fact, to implement our proposed APIs, techniques from probabilistic programming might be useful. Similarly, \cite{sampson2011enerj} propose a programming interface for approximate computation. 

Similar in spirit to our approach is \citep{Kraska2018TheCF} which proposes to incorporate neural models into database systems by replacing existing index structures with neural models that can be both faster and smaller. 
In contrast, we aim not to replace existing data structures or algorithms but transparently integrate with standard algorithms and systems. Our approach is general enough to be used to improve the heuristics in algorithms (as done here), to optimize database systems (similar to \cite{Kraska2018TheCF}), or to simply replace an arbitrarily chosen constant.
Another approach that is similar to \SmartChoices is Spiral \citep{facebook-spiral-2018} but it is far more limited in scope than \SmartChoices in that it aims to predict boolean values only and relies on ground truth data for model building. 

Similarly, a number of papers apply machine learning to algorithmic problems, e.g.\ Neural Turing Machines \citep{Graves2014NeuralTM} aims to build a full neural model for program execution.
\cite{Kaempfer2018LearningTM,Kool2018AttentionSY,Bello2016NeuralCO} propose end-to-end ML approaches to combinatorial optimization problems. In contrast to our approach these approaches \emph{replace} the existing methods with an ML-system rather than augmenting them. These are a good demonstration of the inversion of control problem mentioned above: using ML requires to give full control to the ML system.

There are a few approaches that are related to our use of the initial function, however most common problems where RL is applied do not have a good initial function. 
Generally related is the idea of imitation learning \citep{imitationlearningsurvey} where the agent aims to replicate the behavior of an expert. Typically the amount of training data created by an expert is very limited. Based on imitation learning is the idea to use previously trained agents to kickstart the learning of a new model \citep{Schmitt2018KickstartingDR} where the authors concurrently use a teacher and a student model and encourage the student model to learn from the teacher through an auxiliary loss that is decreased over time as the student becomes better.

In some applications it may be possible to obtain additional training data from experts from other sources, e.g.\ \citep{hester:aaai18, aytar:nips18-youtube} leverage YouTube videos of gameplay to increase training speed of their agents. These approaches work well in cases where it is possible to leverage external data sources.

Caches are an interesting application area where multiple teams have shown in the past that ML can improve cache performance~\citep{Zhong2018-caches,Lykouris2018-cache,Hashemi-2018-cache,Narayanan:2018-cache,Gramacy2002-cache}. In contrast to our approach,
all ML models are built for task-specific caches, and do not generalize to other tasks.
Algorithm selection has been an approach to apply RL for improving sorting algorithms~\citep{Lagoudakis:2000:ASU:645529.657981}.
Search algorithms have also been improved using genetic algorithms to tweak code optimization~\citep{Li2005-search-genetic}.

\section{Conclusion}
We have introduced a new programming concept called a \SmartChoice aiming to make it easier for developers to use machine learning from their existing code in new application areas.
Contrary to other approaches, \SmartChoices can easily be integrated and hand full control to the developer over how ML models are used and trained. Our approach bridge the chasm between the traditional approaches of software systems building and machine learning modeling, and thus allow for the developer to focus on refining their algorithm and metrics rather than working on building pipelines to incorporate machine learning.
We achieve this by proposing a new object called \SmartChoice which provides a 3-call API. A \SmartChoice observes information about its context and receives feedback about the quality of predictions instead of being assigned a value directly.

We have studied the feasibility of \SmartChoices in three algorithmic problems. For each we show how easy \SmartChoices can be incorporated and how performance improves in comparison to not using a \SmartChoice at all. Specifically, through our experiments we highlight both advantages and disadvantages that reinforcement learning brings when used as a solution for a generic interface as \SmartChoices.

Note that we do \emph{not} claim to have the best possible machine learning model for each of these problems but our contribution lies in building a framework that allows for using ML easily, spreading its use, and improving the performance in places where machine learning would not have been used otherwise. \SmartChoices are applicable to more general problems across a large variety of domains from system optimization to user modelling.
Our current implementation of \SmartChoices is built on standard RL methods but other ML methods such as supervised learning are in scope as well if the problem is appropriate.

\paragraph*{Future Work.}
In this paper we barely scratch the surface of the new opportunities created with \SmartChoices. The current rate of progress in ML will enable better results and wider applicability of \SmartChoices to new applications. We hope that \SmartChoices will inspire the use of ML in places where it has not been considered before.

\paragraph*{Acknowledgements.}
The authors are part of a larger effort aiming to hybridize machine learning and programming. We would like to thank all other members of the team for their contributions to this work: 
George Baggott, 
Gabor Bartok, 
Jesse Berent, 
Eugene Brevdo, 
Andrew Bunner, 
Jeff Dean, 
Arkady Epshteyn, 
Sanjay Ghemawat, 
Daniel Golovin, 
Alex Grubb,
Ramki Gummadi, 
Wei Huang,
Eugene Kirpichov, 
Effrosyni Kokiopoulou, 
Ketan Mandke, 
Luciano Sbaiz, 
Benjamin Solnik, 
Weikang Zhou.

Further we would like to thank the authors and contributors of the TF-agents \cite{TFAgents} library: 
Sergio Guadarrama, 
Julian Ibarz, 
Anoop Korattikara, 
Oscar Ramirez.

\small
\bibliography{smartchoices}
\bibliographystyle{icml2019}

\end{document}